\documentclass{ecai} 
\usepackage{latexsym}
\usepackage{amssymb}
\usepackage{amsmath}
\usepackage{amsthm}
\usepackage{booktabs}
\usepackage{enumitem}
\usepackage{graphicx}
\usepackage{color}
\usepackage{algorithm}  
\usepackage{algorithmic}  
\usepackage{multirow}
\usepackage{float}
\usepackage{placeins}  
\usepackage{flushend}  
\usepackage{appendix}

\usepackage{subfigure}
\newcommand{\BibTeX}{B\kern-.05em{\sc i\kern-.025em b}\kern-.08em\TeX}

\begin{document}

%%%%%%%%%%%%%%%%%%%%%%%%%%%%%%%%%%%%%%%%%%%%%%%%%%%%%%%%%%%%%%%%%%%%%%%%

\begin{frontmatter}

%%% Use this command to specify your submission number.
%%% In doubleblind mode, it will be printed on the first page.

\paperid{853}

\title{Talos: A More Effective and Efficient Adversarial Defense for GNN Models Based on the Global Homophily of Graphs}

\author{\fnms{Duanyu}~\snm{Li}} 
\author{\fnms{Huijun}~\snm{Wu}\!\!\! \thanks{Corresponding authors: Huijun Wu (\texttt{wuhuijun@nudt.edu.cn}) and Min Xie (\texttt{xiemin@nudt.edu.cn})}} %%\orcid{0000-0002-9513-5359}
\author{\fnms{Min}~\snm{Xie}\!\!\! \footnotemark[*]}
\author{\fnms{Xugang}~\snm{Wu} } 
\author{\fnms{Zhenwei}~\snm{Wu}} 
\author{\fnms{Wenzhe}~\snm{Zhang}}

\address{School of Computer Science,National University of Defense Technology}

\begin{abstract}
Graph neural network (GNN) models play a pivotal role in numerous tasks involving graph-related data analysis. Despite their efficacy, similar to other deep learning models, GNNs are susceptible to adversarial attacks. Even minor perturbations in graph data can induce substantial alterations in model predictions. While existing research has explored various adversarial defense techniques for GNNs, the challenge of defending against adversarial attacks on real-world scale graph data remains largely unresolved. On one hand, methods reliant on graph purification and preprocessing tend to excessively emphasize local graph information, leading to sub-optimal defensive outcomes. On the other hand, approaches rooted in graph structure learning entail significant time overheads, rendering them impractical for large-scale graphs. In this paper, we propose a new defense method named Talos, which enhances the global, rather than local, homophily of graphs as a defense. Experiments show that the proposed approach notably outperforms state-of-the-art defense approaches, while imposing little computational overhead.

\end{abstract}

\end{frontmatter}

%%%%%%%%%%%%%%%%%%%%%%%%%%%%%%%%%%%%%%%%%%%%%%%%%%%%%%%%%%%%%%%%%%%%%%%%

\section{Introduction}
%我们是否可以下结论，现有的GNN对抗要么是基于图结构学习，要么是基于预处理
Graph Neural Networks (GNNs) extend deep learning to graph data and significantly outperform traditional data analytic methods in tasks such as node classification\citep{hamilton2017inductive}, graph classification \citep{wang2021mixup}, and link prediction\citep{zhang2018link}. However, similar to deep learning models for computer vision~\citep{goodfellow2014explaining} and natural language processing~\citep{chang2021robustness}, graph neural network models also suffer from adversarial attacks, where adversaries craft subtle perturbations to mislead the model into making incorrect predictions\citep{gunnemann2022graph,xu2021robustness,zhang2024trustworthy}. The lack of adversarial robustness makes it hard to apply GNN models to critical real-world scenarios. 

To address this challenge, recent studies propose various methods to defend against adversarial attacks. Among them, many leverage the concept of homophily to design their defense mechanisms. Recent research indicates that GNN models fundamentally aggregate similar neighbors to enhance node representations\citep{GNNJaccard,zhu2020beyond} so that they can more effectively utilize the graph structure information that was previously overlooked. Adversarial attacks, therefore, aim to disrupt homophily by adding dissimilar edges or altering node features \cite{GNNJaccard,jin2020graph,tao2021single}. Defense methods utilize the homophily assumption differently. For instance, GCN-Jaccard directly purifies the input graph by removing edges connecting dissimilar nodes. Soft-Median/Medoid\cite{SoftMedian,geisler2020reliable} aggregate the approximated median and medoid of neighboring features to mitigate the impact of adversarially added neighbors, which are often extremely different from the central nodes. Pro-GNN\citep{jin2020graph} implicitly incorporates the constraint of graph homophily in its model architecture. The loss function of the Pro-GNN model punishes the edges which connect dissimilar nodes.

Although these defense methods show promising results in mitigating the impact of attacks, a recent study shows they can be easily subjected to adaptive attacks\citep{mujkanovic2022defenses}. In this paper, we argue that one important reason is that they focus solely on the local neighborhood of nodes but overlook the fact that graph homophily is a global concept that involves multi-hop neighbors of nodes. To comprehensively enhance the robustness of GNN models, we should reconsider how to improve the overall homophily of graphs. Meanwhile,  real-world graphs consist of huge amounts of nodes and edges, and many of the existing defenses introduce significant overheads and make them hard to use in practice. For example, the time overhead of GNNGuard\cite{GNNGuard} grows with the number of node features, making it ineffective for processing real-world graph data. GCN-SVD\cite{GCNSVD} and ProGNN\cite{jin2020graph} rely on expensive SVD decomposition to achieve defense. We believe that efficiency should be prioritized for the defense design.

In this paper, we propose a more effective and efficient adversarial defense method, Talos, that enhances the global, rather than local, homophily of input graphs to defend against adversarial attacks. Talos preprocesses the graph only once and can be computed efficiently. Our contributions are outlined as follows:
\begin{itemize}  

\item We have expanded the scope of defense to encompass neighbors spanning arbitrary hops across the entire graph, thereby ensuring comprehensive and robust protection.
\item Theoretical derivations are provided for our strategy, enhancing its theoretical validity and effectiveness.
\item Experiments show that Talos is not only more effective but also much faster compared with state-of-the-art defense methods. 
\end{itemize}

The rest of this paper is organized as follows: Section 2 introduces the related work, and Section 3 presents the methodology. We show experimental results in Section 4 and conclude in Section 5.

%%%%%%%%%%%%%%%%%%%%%%%%%%%%%%%%%%%%%%%%%%%%%%%%%%%%%%%%%%%%%%%%%%%%%%%%
\section{Related Work}
% 这里就是对比一下现有工作，一方面强调我们提出的方法复杂度低（一些基于学习的方法也考虑到了全局信息，但是这些方法复杂度比较高，比如要想进行训练就不得不依赖强大的计算资源，而我们的方法memory footprint很低），另一方面强调我们的方法效果好。
\subsection{Graph Adversarial Attacks} 
%总体思路：首先介绍一下图上面的对抗攻防和image上面的攻防有一些不同，比如离散，比如transductive，为了解决这些问题。nettack采用线性近似来构建对抗扰动，IG系列通过集成梯度来构建攻击。不过这些攻击主要还是局部攻击，针对个别节点进行攻击。针对全局攻击

Due to the discrete features and transductive learning settings, generating adversarial perturbations for GNN models exhibits unique challenges. To tackle these problems, Dai et al.\citep{dai2018adversarial} use reinforcement learning to craft generalized attacks. Nettack\citep{Nettack} uses a linear model, which removes the non-linear components of GNN models, as a surrogate to solve the problem. IG-FSGM and IG-JSMA\citep{GNNJaccard} rely on integrated gradients to accurately estimate gradients in discrete states. This allows for a precise measurement of the impact of perturbations on model outputs. Additionally, PGDAttack\citep{PGD} optimizes attack strategies using gradient descent algorithms. It disrupts the graph representation learning process of GNNs by manipulating the edges. The above-mentioned approaches, however, primarily focus on local attacks that target specific nodes. Metaattack\citep{Metattack} utilizes meta-learning techniques to carry out poisoning attacks during the training process of graph data. This results in reduced accuracy in tasks such as classification and clustering. Most of the existing attack methods are gradient-based, making it hard to be applied to large-scale graphs. PR-BCD\citep{SoftMedian} proposes a sparsity-aware attack method that adds and removes edges between existing nodes based on Randomized Block Coordinate Descent (R-BCD).

\subsection{Graph Adversarial Defense and Challenges}

To improve the robustness of GNN models, defense methods are actively explored. Wu et al.\citep{GNNJaccard} found that perturbing the graph structure has a more significant impact than tampering with the features. Attacks tend to add edges rather than delete them. They also revealed that edges are mostly added between dissimilar nodes. Based on such observations, they proposed GCN-Jaccard, which utilizes Jaccard similarities to identify and eliminate the perturbed edges. This effectively purifies the graph data and enhances its robustness against attacks. Some other methods use a similar idea to filter out dissimilar edges. For example, SoftMedian \citep{SoftMedian} and SoftMedoid\citep{geisler2020reliable} take advantage of recent advancements in differentiable sorting to design robust aggregation functions in GNN models. As mid-frequency signals partly correspond to moderately similar edges, Mid-GCN\citep{{huang2023robust}} preserves the mid-frequency signals while abandoning the high/low-frequency signals to defend against adversarial attacks. GNNGuard\citep{GNNGuard} employs an attention mechanism to assign higher weights to edges between similar nodes. GRAND\citep{feng2020graph} uses random feature augmentations together with neighborhood augmentation to avoid the over-reliance on directly connected neighbors. However, the above methods focus solely on the local neighborhood of nodes, neglecting to consider the overall structure of graphs. This could potentially expose them to adaptive attacks by adversaries\citep{mujkanovic2022defenses}.as attackers can allocate more attacking budget to further neighbors of nodes to achieve adaptive attacks.

In addition to removing adversarial direct neighbors, some other defense methods propose to modify the model architecture. GCN-SVD\citep{GCNSVD} obtains a purified graph by replacing the perturbed graph with its low-rank approximation. Pro-GNN\citep{jin2020graph} introduces a regularization term during the training process, guiding the model to generate a low-rank and sparse graph. Since most adversarial attacks focus on manipulating the graph structure, Wu et al.\citep{wu2022towards}  propose to use a co-training mechanism, which employs different models to fully utilize both the information of node features and graph structures. Unfortunately, these methods often involve time-consuming processes that hinder effective processing. Consequently, they struggle to efficiently handle larger and more complex real-world graph data. Moreover, existing adversarial defense methods heavily rely on empirical observations but lack theoretical analysis, making them less reliable for real-world scenarios.

In this paper, we propose Talos, which takes the entire graph structure into account. Talos preprocesses the graph by enhancing its homophily and leaves the model training unchanged, making it more efficient and applicable to large-scale graphs.

\section{Methodology}
%学术写作建议遵循总分的原则，也就是每一节的开始实际上是对整节内容的一个凝练和总结。所以这里对于后面方法的主要思想应该要提及一下。
This section presents the design of Talos. The fundamental principle of Talos involves harnessing global homophily information to promptly identify the edge most likely added by the attack and eliminate it. We will provide a comprehensive explanation and derivation of the Talos approach.

\subsection{Notations}
Let  $\mathcal{G}=(\mathcal{V}, \mathcal{E}, \mathbf{X})$ be a graph, where $\mathcal{V}$ represents the set of nodes containing $n$ nodes $\left\{v_{1}, \cdots, v_{n}\right\}$ with $|\mathcal{V}|=n$. $\mathcal{E}$ denotes the set of edges, typically represented by an adjacency matrix $\mathbf{A} \in \{0,1\}^{n \times n}$. An entry $\mathbf{A}_{ij} = 1$ indicates the presence of an edge $(v_{i}, v_{j}) \in \mathcal{E}$, while 0 indicates its absence.

Additionally, we have the feature matrix $\mathbf{X}$, where the features of each node $v \in \mathcal{V}$ are represented as a d-dimensional feature vector $\mathbf{x}_{v} \in \mathbb{R}^{d}$. The feature matrix $\mathbf{X}$ comprises these feature vectors, specifically $\mathbf{X}=\left(\mathbf{x}_{1}, \mathbf{x}_{2}, \cdots, \mathbf{x}_{n}\right)^{\mathrm{T}} \in \mathbb{R}^{n \times m}$, where $m$ denotes the number of features.

In the context of graph node classification tasks, each node is associated with a label $y$, and $\mathcal{Y}^{|\mathcal{V}|}$ represents the set of true labels for all nodes. The goal of Talos is to enhance the model's prediction capability after an attack.

\subsection{Effectiveness of High-order Neighbor Information }

\begin{figure}[ht]
    \centering
    % \vspace{-10pt}
    \includegraphics[width=0.53\linewidth]{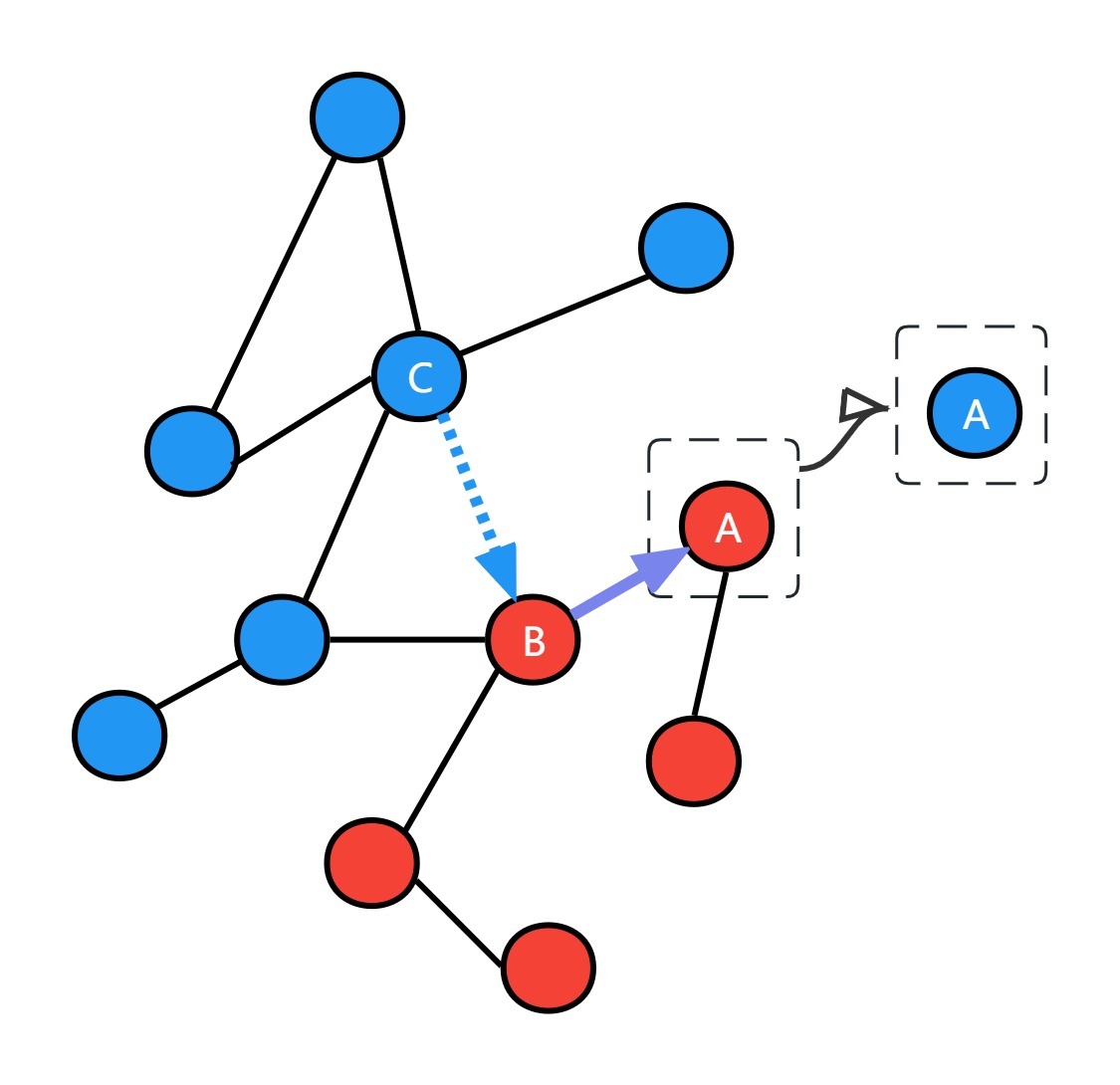}
    \caption{Indirect Attack}
    \vspace{10pt}
    \label{fig:2orderAttack}
\end{figure}

Existing adversarial attacks, whether approximation-based or gradient-based, allow perturbations to arbitrary edges within the graph. Existing defenses like GCN-Jaccard, however, often focus solely on the local neighborhood information. This approach leaves them susceptible to adaptive adversarial attacks designed to exploit these defenses. As depicted in Figure~\ref{fig:2orderAttack}, an attacker can meticulously orchestrate a perturbation attack on node A while circumventing direct edge manipulations. By connecting nodes B and C, the attacker indirectly contaminates node B and triggers misclassification of A during information propagation. Nonetheless, if the defensive strategy only considers A's first-order neighborhood information and employs GCN-Jaccard for defense, it may prove ineffective.

\begin{figure}
    \centering
    \includegraphics[width=0.8\linewidth]{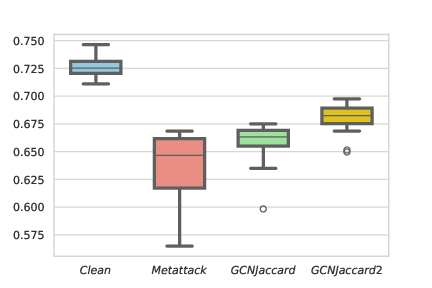}
    \caption{The augmentation experiment on the GCN-Jaccard model}
    \vspace{10pt}
    \label{fig:2orderDefense}
\end{figure}

From this, we can deduce the importance of considering global information during defensive strategies. To substantiate this claim, we conducted a fundamental augmentation experiment on the GCN-Jaccard model by disconnecting all dissimilar second-order neighbor node pairs. The process is as follows:

By squaring the adjacency matrix, we identify all second-order neighbor node pairs $(v_i, v_j)$ that satisfy $\mathbf{A}^2[i,j]>0$ and calculate their similarity using the Jaccard coefficient. Subsequently, we select the node pairs with lower similarity and traverse all intermediate nodes (denoted as $v_k$). In each traversal, we remove the edge $e_{ik}$ or $e_{jk}$ with the lower similarity, effectively pruning all dissimilar second-order adjacent node pairs in a direct and intuitive manner. With this technique, we evaluate the accuracy of GCN models on the Citeseer dataset. As shown in Figure~\ref{fig:2orderDefense}, utilizing higher-order neighbor information leads to significant potential for enhancing the overall defensive capability of the model.

% \vspace{-10pt}

\subsection{Global Homophily Index}

Despite the potential of improved robustness, high-order modifications to GCN-Jaccard may remove too many edges, thus leading to sub-optimal accuracy. For example, on the Cora dataset, setting the Jaccard threshold to 0.01 results in 548 first-order dissimilar node pairs, while the number of third-order dissimilar node pairs increases dramatically to 44,799 pairs. This suggests that removing connections between third-order dissimilar node pairs could entail the loss of a substantial number of edges. To address this issue, we formulate a global homophily index that quantifies the amount of information transmitted by all edges in the entire graph across all orders. By comparing the change in the global homophily index before and after the removal of an edge, we can evaluate the edge's importance in information transmission from low to high orders.

Consider the node pair $(v_i, v_j)$, which may have multiple intermediate nodes between them. 
In GNN models, the homophily information between these nodes is conveyed through the graph aggregation mechanism (where Jaccard similarity is used as the metric for measuring homophily). However, during the information transmission process, the message-passing mechanism can cause the loss of homophily information, which will be further strengthened by the number of paths. Consequently, our model construction takes into account both the \textbf{number of paths} and the \textbf{rate of information loss}.

\subsubsection{The Number of Paths}

We first consider how to obtain the number of paths $n$. Taking the $k$th power of the adjacency matrix can provide the number of $k$-hop paths between node pairs, which can be represented as $ \mathbf{A}^k_{ij}$. Therefore, we can model the $k$th-order global homophily index $\mathrm{Hom}^{(k)}$ as follows:
\begin{equation}    
\label{equ:1}
\mathrm{Hom}^{(k)}= <\mathbf{A}^k,\mathcal{J}>
\end{equation}  
where $\mathbf{A}$ denotes the adjacency matrix, $\mathcal{J}$ is the Jaccard similarity matrix, and $<\cdot,\cdot>$ signifies the operation of matrix inner product.

\subsubsection{The Rate of Information Loss}

Zhu et al. \citep{DBLP:conf/nips/ZhuYZHAK20}formally define the representation learned by a GNN for each node $v$ at the $k$th iteration as follows:

\begin{equation}    
\label{equ:gnn}
\mathbf{r}_{v}^{(k)} = \text{COMBINE}\left(\mathbf{r}_{v}^{(k-1)}, \text{AGGR}\left(\left\{\mathbf{r}_{u}^{(k-1)}: u \in \bar{N}(v)\right\}\right)\right)  
\end{equation}  

Different GNN designs show distinct AGGR() and COMBINE() functions. AGGR() (e.g., Mean) computes the aggregation of neighbor embeddings, while COMBINE() integrates the self-embedding with the aggregated neighbor embeddings. The design of these two functions invariably introduces information loss. Hence, we denote the mean retention rate of the residual information following information aggregation as $\alpha$ (where $\alpha \in [0,1]$). Subsequently, the global homophily index can be formulated as follows:
 
\begin{equation}  
\label{equ:2}
\mathrm{Hom}=\sum_{k=0}^{\infty} \alpha^{k} \mathrm{Hom}^{(k)}
\end{equation}  

After deriving Formulas~\ref{equ:1} and~\ref{equ:2}, we can calculate $\mathrm{Hom}$ as follows.

\begin{equation}  
\label{equ:3}
\mathrm{Hom}=<(I-\alpha\mathbf{A})^{-1},\mathcal{J}>
\end{equation}  

For the equation $\sum_{k=0}^{\infty} \alpha^k \mathbf{A}^k = (I - \alpha \mathbf{A})^{-1}$, certain conditions be met. Specifically, the convergence parameter $\alpha$ must satisfy the inequality $\alpha < \frac{1}{\rho(\mathbf{A})}$, where $\rho(\mathbf{A})$ denotes the spectral radius of the adjacency matrix $\mathbf{A}$. To simplify the subsequent exposition, we employ the unified notation $M = (I-\alpha\mathbf{A})^{-1}$.

\subsection{Derivation and Analysis}
We model the global homophily index to identify and eliminate edges that propagate the most misleading information.  That is, we aim to find an edge whose removal would maximize $\Delta \mathrm{Hom}$. This can be mathematically formulated as:
\begin{equation}  
\label{equ:4}
\underset{\Delta \mathbf{A}}{\mathrm{argmax}} \Delta \mathrm{Hom}
\end{equation}

Suppose the edge $ e_{kl} $ is removed, resulting in a new adjacency matrix $\mathbf{A}^{\prime}$. Here, $\Delta \mathbf{A} = \mathbf{A}^{\prime} - \mathbf{A}$. For $\Delta \mathbf{A}$, $\Delta \mathbf{A}_{kl} = \Delta \mathbf{A}_{lk} = -1$, with all other values being 0. We further derived and optimized this result to achieve the final outcome.

\begin{equation}  
\label{equ:5}
% \begin{aligned}  
\Delta \mathrm{Hom} = <M^{\prime} - M, \mathcal{J}> = \alpha <M^{\prime} \Delta \mathbf{A} M, \mathcal{J}>  
% \end{aligned}  
\end{equation}

Where $M^{\prime} = (I - \alpha\mathbf{A}^{\prime})^{-1}$. Given this modification, we can determine the values at each position in the matrix $M^{\prime}\Delta \mathbf{A} M$. The calculation process is detailed as follows:

\begin{equation}  
\label{equ:6}
% \begin{aligned}  
{\left[M^{\prime}\Delta \mathbf{A} M\right]_{i j}} = \sum_{s=1}^{n} \sum_{t=1}^{n} M_{i s}^{\prime} \Delta A_{s t} M_{t j} = -M_{i l}^{\prime} M_{k j} - M_{i k}^{\prime} M_{l j}  
% \end{aligned}  
\end{equation}

By substituting Equation (6) into Equation (5), we can perform detailed mathematical derivations and obtain the following results\begin{equation}  
\label{equ:7}
\begin{aligned}  
\Delta \mathrm{Hom} &= \alpha < M^{\prime}\Delta \mathbf{A} M , \mathcal{J}> = \alpha \sum_{i, j}\left[M^{\prime}\Delta \mathbf{A} M\right]_{i j} \mathcal{J}_{i j} \\  
&= -\alpha \sum_{i, j}(M_{i l}^{\prime} M_{k j}+M_{i k}^{\prime} M_{l j})\mathcal{J}_{i j} \\  
&= -\alpha \sum_{i, j}(M_{l i}^{\prime} \mathcal{J}_{i j} M_{j k}+M_{k i}^{\prime} \mathcal{J}_{i j} M_{j l}) \\  
% &= -\alpha\left[M^{\prime} \mathcal{J} M\right]_{l k}-\alpha\left[M^{\prime} \mathcal{J} M\right]_{k l} \\  
&= -2 \alpha\left[M^{\prime} \mathcal{J} M\right]_{k l}  
\end{aligned}  
\end{equation}

Further derivation leads to a conclusion.\begin{equation}  
\label{equ:8}
\underset{\Delta \mathbf{A}}{\mathrm{argmax}} \Delta  \mathrm{Hom} = \underset{\Delta \mathbf{A}}{\mathrm{argmin}}<M^{\prime} \mathcal{J} M,-\Delta \mathbf{A}>
\end{equation}

%通过上述的数学推导和分析，我们对去除某条边后对全局同配性指数的影响用数学公式进行了表示，为我们 Generalizing Defense Strategy的提出提供了数学支持。在后面的内容中，我们将提出该策略的具体实施方案，并且通过一些举措来使得该策略是轻量化且快捷的。
We have mathematically represented the impact on the global homophily index after the removal of an edge, providing mathematical support for Talos.  In subsequent content, we will present its implementation and adopt various optimizations to ensure that the strategy is lightweight and efficient.

\subsection{Optimizations}

As shown in the previous section, Talos can be formulated as solving the problem of $\mathrm{argmin}_{\Delta \mathbf{A}}<M^{\prime} \mathcal{J} M,-\Delta \mathbf{A}>$. However, there is a notable challenge to implement it.  Whenever the decision is made to remove a specific edge, it necessitates traversing and computing the matrix $M^{\prime} \mathcal{J} M$, which significantly increases the computational complexity. Fortunately, our primary objective is not to obtain precise numerical values but rather to effectively rank and evaluate edges with greater levels of harm. In light of this, we introduce an approximate optimization to the method, which enables substantial enhancement for computational efficiency.

\subsubsection{Approximate Optimization Strategy}

Due to the minor perturbation of state $\mathbf{A}$ before and after defending a specific edge, we reasonably postulate the approximation between $\mathbf{A}$ and $\mathbf{A}'$. Consequently, we can also approximate the similarity between $M'$ and $M$. Grounded in this approximation, we can approximate the original formula as follows:
\begin{equation}
\label{equ:9}
\underset{\Delta \mathbf{A}}{\mathrm{argmax}} \Delta \mathrm{Hom} \approx \underset{\Delta \mathbf{A}}{\mathrm{argmin}} <M \mathcal{J} M, -\Delta \mathbf{A}>  
\end{equation}

This approximation allows us to circumvent the intricate computation involving M', thereby simplifying the computation.

\subsubsection{Efficient Batch Selection Strategy}

If the recalculation of $M \mathcal{J} M$ is required every time an edge is removed, the computational cost becomes prohibitively high. However, in essence, precise calculations are not our primary need; rather, we seek to obtain a ranking of edges based on their homophily. Therefore, we can directly identify the edges corresponding to the positions of the smallest $n$ values in the matrix $M \mathcal{J} M$ and focus our defense efforts on these edges. This approach eliminates the need for performing $n$ iterations and repeatedly computing a new $M \mathcal{J} M$ to select the minimum value in each iteration.

\subsubsection{Matrix Computation Strategy}
Considering that node features in graph datasets are often discrete, as exemplified by common graph datasets such as CORA where the feature matrix X contains elements with values limited to {0,1}. Given this context, we can compute the Jaccard coefficient between node $v_{i}$ and node $v_{j}$ using the following approach:
\begin{equation}
\label{equ:10}
\mathcal{J}_{ij}=\frac{\mathbf{x}_{i} \mathbf{x}_{j}^{T}}{1-\left(1-\mathbf{x}_{i}\right)\left(1-\mathbf{x}_{j}\right)^{T}}=\frac{\mathbf{x}_{i} \mathbf{x}_{j}^{T}}{1-\bar{\mathbf{x}}_{i} \bar{\mathbf{x}}_{j}^{T}}=\frac{\mathbf{x}_{i} \mathbf{x}_{j}^{T}}{\overline{\bar{\mathbf{x}}_{i} \bar{\mathbf{x}}_{j}^{T}}} 
\end{equation}
This further leads to $\mathcal{J}=\frac{X X^{T}}{\overline{\bar{X} \bar{X}^{T}}}$
% \begin{equation}
% \label{equ:11}
% \end{equation}

By employing matrix operations, we can achieve significant computational savings, as the division operation here refers to element-wise division between corresponding positions in the matrices.

\begin{algorithm}
\caption{Defense procedure of Talos}  
\vspace{10pt}
\label{alg:defense_strategy}  
\begin{algorithmic}[1]  
\REQUIRE Poisoned graph $\mathcal{G}_p = (\mathcal{V}, \mathcal{E}_p, \mathbf{X})$ with adjacency matrix $\mathbf{\mathbf{A}}_p$; Retention Rate $\alpha$; Number of edges to cut $n$  
  
\STATE Compute the Jaccard similarity matrix $\mathcal{J}$ using Eq.~(\ref{equ:10});  
\STATE Evaluate $M\mathcal{J}M$, where $M = (1 - \mathbf{\mathbf{A}}_p)^{-1}$;  
\STATE Identify the $n$ edge zeroes in $\mathbf{\mathbf{A}}_p$ corresponding to the smallest values in $M\mathcal{J}M$, resulting in matrix $\mathbf{A}_c$;  
\STATE Purifying the edge set of the graph structure into $\mathcal{E}_c$ based on $\mathbf{A}_c$, we obtain the sanitized clean graph $\mathcal{G}_c = (\mathcal{V}, \mathcal{E}_c, \mathbf{X})$;
\RETURN $\mathcal{G}_c$  
\end{algorithmic}  

\end{algorithm}
% \vspace{-15pt}

% \begin{table*}[htbp]
%     \centering
%     % \FloatBarrier  
%     % \renewcommand{\arraystretch}{1.3} % Adjust line height for readability
%     \caption{Attack Result}
%     \label{tab:2}
%     \begin{tabular}{@{}l c cccccc@{}}
%     \toprule
%     \multirow{2}{*}{\textbf{Dataset}} & \multirow{2}{*}{\textbf{Attack}} & \multicolumn{6}{c}{\textbf{Perturbation Rate}} \\
%     \cmidrule{3-8}
%     & & \textbf{0.00} & \textbf{0.05} & \textbf{0.10} & \textbf{0.15} & \textbf{0.20} & \textbf{0.25} \\
%     \midrule
%     \multirow{2}{*}{Cora}
%     & Metattack &83.11±0.41&79.38±0.72&75.31±0.50&64.16±1.00&55.23±1.01&52.44±0.39 \\
%     & PGD &82.87±0.45&78.74±0.21&75.80±0.77&74.78±0.53&73.01±0.63&71.86±0.58 \\

%     \midrule
%     \multirow{2}{*}{Photo}
%     & Metattack & 93.80±0.21&90.19±1.11&84.18±1.06&73.84±2.83&59.54±0.43&57.65±1.57 \\
%     & PGD &93.64±0.25&85.23±0.21&82.06±0.30&79.56±0.70&78.42±0.45&76.46±1.03 \\
    
%     \midrule
%     \multirow{2}{*}{Computers}
%     & Metattack & 88.30±0.43 & 81.68±1.06 & 74.20±1.84 & 72.10±1.57 & 71.06±1.40 & 69.50±1.16  \\
%     & PGD & 88.30±0.43 & 79.22±0.56 &  74.40±0.74 & 70.72±0.69 & 68.58±0.89 & 66.66±0.88 \\

%     \bottomrule
%     \end{tabular}
%     \end{table*}

\subsection{Overview of Talos}

As shown in Algorithm~\ref{alg:defense_strategy}, Talos can be seamlessly incorporated into existing GNNs. Given a graph $\mathcal{G}_p = (\mathcal{V}, \mathcal{E}_p, \mathbf{X})$ subjected to poisoning attacks, we can efficiently identify and remove suspected attacked edges, thus obtaining a clean graph structure data $\mathcal{G}_c$. Talos first calculates the Jaccard similarity matrix with reference to Eq.~(\ref{equ:10}). Then it computes the similar gradient matrix $M\mathcal{J}M$, and we only need to select the smallest $n$ values in the lower triangular part of this symmetric matrix and delete the corresponding edges.

% \begin{table}[h]
% \centering  
% \caption{Dataset Summary}  
% \label{tab:1}  
% \begin{tabular}{lrrrr}  
% \toprule  

% \textbf{Dataset} & \textbf{Nodes} & \textbf{Edges} & \textbf{Classes} & \textbf{Features} \\ \midrule  
% Cora             & 2,708          & 5,429          & 7                & 1,433               \\  
% Photo            & 7,487          & 119,043        & 8                & 745                 \\ 
% Computers        & 13,381         & 245,778        & 10               & 767                  \\ 
% \bottomrule  

% \end{tabular}  
% \end{table}

\section{Experiments}
In this section, we assess the efficacy of Talos against non-targeted adversarial attacks on graphs. Experiments were conducted on servers equipped with NVIDIA Tesla V100-SXM2 32GB GPUs. Through the experiments, we evaluate how Talos performs compared with the state-of-the-art defense methods. We also show that Talos can be applied to different types of GNN models.

\subsection{Setup}
\subsubsection{Datasets}
In this study, we primarily utilized three widely recognized datasets: Cora\cite{CCGC}, Amazon Photo\cite{CONVERT}, and Amazon Computers. The selection of these three datasets was based on three main considerations:

Firstly, these datasets are commonly used to evaluate the performance of GNN algorithms.

Secondly, their scales cover the needs of small, medium, and large-scale graph-structured data, allowing for a more comprehensive evaluation of the algorithms.

Thirdly, under the constraints of our experimental conditions, these three datasets could undergo a thorough evaluation. As for the widely regarded large dataset, Pubmed, due to experimental constraints, we followed the work of Daniel Zügner~\citep{SoftMedian} and conducted separate tests on it. In the experiments, we operated on the largest connected subgraph for each dataset. All datasets were randomly divided, with 10\% of the data allocated for training, 10\% for validation, and the remaining 80\% for testing, thereby providing a rigorous environment for evaluating model performance.

\subsubsection{Generating Adversarial Attacks}

To evaluate the effectiveness of Talos, we first conducted attacks on graph data. We utilized two global attack methods: Metattack\citep{Metattack} and Projected Gradient Descent (PGD)\citep{PGD}. During the implementation of the Metattack, we employed a simple GCN as a proxy model to guide the direction of the attack. This proxy model was configured with a two-layer structure, including a hidden layer with 64 nodes. To reduce the risk of overfitting and enhance the model's generalization ability, we set a dropout ratio of 0.5. The learning rate (lr) was set to 0.01, and the weight decay was set to 0. Additionally, the model was trained for 1000 epochs with an early stopping strategy to prevent overfitting. For Metattack itself, we utilized a meta-gradient approach with self-training and set the momentum parameter to 0.9. During the implementation of the PGDAttack, we conducted multiple attacks on datasets (Cora, Amazon Photo, and Amazon Computers) using these two adversarial attack methods. In these attacks, we perturbed the edges in the graph, with the proportion of attacked edges ranging from 5\% to 25\% in steps of 5\%. After the attacks, we used a GCN with the same settings as the Metattack proxy model for detection to evaluate the performance of the graph data after poisoning attacks. We have detailed the results of the attacks in Table~\ref{tab:2}, and the attacked graph data will serve as the foundation for further defense evaluation.
    
\begin{table}[ht]
    \centering
    % \FloatBarrier  
    % \renewcommand{\arraystretch}{1.3} % Adjust line height for readability
    \caption{Attack Result}
    \vspace{10pt}
    \label{tab:2}
    \begin{tabular}{cccc}
    \toprule
    \textbf{Dataset} & \textbf{Ptb Rate(\%)} & \textbf{Metattack(\%)} & \textbf{PGD(\%)} \\
    \midrule
    \multirow{5}{*}{\textbf{Cora}}
    % & 0 & 83.11±0.41 & 82.87±0.45 \\
    & 5 & 79.38±0.72 & 78.74±0.21 \\
    & 10 & 75.31±0.50 & 75.80±0.77 \\
    & 15 & 64.16±1.00 & 74.78±0.53 \\
    & 20 & 55.23±1.01 & 73.01±0.63 \\
    & 25 & 52.44±0.39 & 71.86±0.58 \\
    
    \midrule
    \multirow{5}{*}{\textbf{Photo}}
    % & 0 & 93.80±0.21 & 93.64±0.25 \\
    & 5 & 90.19±1.11 & 85.23±0.21 \\
    & 10 & 84.18±1.06 & 82.06±0.30 \\
    & 15 & 73.84±2.83 & 79.56±0.70 \\
    & 20 & 59.54±0.43 & 78.42±0.45 \\
    & 25 & 57.65±1.57 & 76.46±1.03 \\
    
    \midrule
    \multirow{5}{*}{\textbf{Computers}}
    % & 0 & 88.30±0.43 & 88.30±0.43 \\
    & 5 & 81.68±1.06 & 79.22±0.56 \\
    & 10 & 74.20±1.84 & 74.40±0.74 \\
    & 15 & 72.10±1.57 & 70.72±0.69 \\
    & 20 & 71.06±1.40 & 68.58±0.89 \\
    & 25 & 69.50±1.16 & 66.66±0.88 \\
    
    \bottomrule
    \end{tabular}
    \end{table}
    % \vspace{-10pt}

\begin{table*}[ht]
    \centering
    \footnotesize
    \caption{Node Classification Performance under Metattack and PGD(\%)}
    \vspace{10pt}
    \renewcommand{\arraystretch}{1.1}
    \label{Performance under Metattack}
    \begin{tabular}{ccc|c|ccccccc}
    \toprule
    DataSet & Attack & Ptb Rate(\%) & GCN & GCN-Jaccard & GCN-SVD & RobustGCN & GNNGuard & Pro-GNN & Talos \\
    \midrule
    % Metattack
    \multirow{12}{*}{Cora} 
     
     & \multirow{6}{*}{Metattack} 
       & 0 & 83.11±0.41 & 80.89±0.10$^4$ & 76.51±0.52$^6$ & 81.49±0.33$^3$ & 76.62±1.24$^5$ & 81.66±2.50$^2$ & \textbf{82.71±0.49$^1$} \\
     & & 5 & 79.38±0.72 & 79.31±0.45$^2$ & 75.11±0.57$^5$ & 77.65±0.83$^4$ & 75.00±0.45$^6$ & 79.08±1.60$^3$ & \textbf{80.83±0.74$^1$} \\
     & & 10 & 75.31±0.50 & 77.08±0.07$^2$ & 72.34±0.34$^6$ & 73.37±1.12$^5$ & 74.84±0.51$^4$ & 76.64±2.15$^3$ & \textbf{78.31±0.76$^1$} \\
     & & 15 & 64.16±1.00 & 72.90±0.38$^2$ & 66.05±0.33$^5$ & 63.61±1.31$^6$ & 71.00±1.65$^4$ & 72.21±1.53$^3$ & \textbf{74.17±0.39$^1$} \\
     & & 20 & 55.23±1.01 & 65.88±1.39$^3$ & 54.09±0.81$^6$ & 55.31±1.22$^5$ & 65.98±2.57$^2$ & 58.47±1.12$^4$ & \textbf{68.19±0.98$^1$} \\
     & & 25 & 52.44±0.39 & 60.96±0.68$^3$ & 50.01±0.79$^6$ & 50.60±0.46$^5$ & \textbf{65.81±1.78$^1$} & 53.12±0.76$^4$ & 65.28±1.16$^2$ \\
     \cmidrule{2-10}
     &\multirow{6}{*}{PGD} 
       & 0 & 82.87±0.45 & 80.94±0.28$^4$ & 75.60±0.48$^6$ & \textbf{81.95±0.23$^1$} & 76.81±0.36$^5$ & 81.16±2.17$^2$ & 81.06±0.91$^3$ \\
     & & 5 & 78.74±0.21 & \textbf{79.03±0.43$^1$} & 76.73±0.71$^5$ & 77.40±0.46$^4$ & 76.07±0.61$^6$ & 78.52±1.38$^3$ & 78.81±0.47$^2$ \\
     & & 10 & 75.80±0.77 & 76.24±0.30$^2$ & 76.07±0.44$^3$ & 75.90±0.84$^4$ & 75.30±0.62$^6$ & 75.57±0.48$^5$ & \textbf{78.42±0.54$^1$} \\
     & & 15 & 74.78±0.53 & 74.56±0.73$^4$ & 73.84±1.05$^6$ & 74.78±0.51$^3$ & 75.63±0.74$^2$ & 74.01±0.58$^5$ & \textbf{75.97±0.41$^1$} \\
     & & 20 & 73.01±0.63 & 73.01±0.66$^3$ & 71.77±1.07$^6$ & 72.38±0.73$^5$ & 74.16±0.16$^2$ & 72.39±0.47$^4$ & \textbf{75.40±0.64$^1$} \\
     & & 25 & 71.86±0.58 & 72.39±0.56$^3$ & 72.36±0.13$^4$ & 71.94±0.29$^5$ & 72.71±1.09$^2$ & 71.79±0.11$^6$ & \textbf{73.98±0.34$^1$} \\
     
    \cmidrule{1-10}
    \multirow{12}{*}{Photo}      
    
    & \multirow{6}{*}{Metattack} 
     & 0 & 93.80±0.21 & 93.09±0.07$^4$ & 90.82±0.15$^5$ & 93.35±0.21$^2$ & 93.20±0.44$^3$ & 90.36±1.39$^6$ & \textbf{93.60±0.27$^1$} \\
    & & 5 & 90.19±1.11 & \textbf{92.29±0.22$^1$} & 87.73±0.24$^5$ & 89.19±1.41$^3$ & 88.70±2.54$^4$ & 87.52±1.36$^6$ & 92.17±0.38$^2$ \\
    & & 10 & 84.18±1.06 & 90.20±0.42$^2$ & 84.00±0.62$^4$ & 84.00±0.35$^5$ & 85.57±0.43$^3$ & 83.78±3.03$^6$ & \textbf{90.34±0.49$^1$} \\
    & & 15 & 73.84±2.83 & 84.28±1.16$^2$ & 68.63±4.44$^6$ & 71.73±3.19$^5$ & 79.28±0.80$^3$ & 77.82±2.04$^4$ & \textbf{88.60±0.70$^1$} \\
    & & 20 & 59.54±0.43 & 82.87±1.52$^2$ & 62.59±3.21$^5$ & 60.53±3.45$^6$ & 66.56±1.99$^4$ & 76.54±2.27$^3$ & \textbf{87.29±0.66$^1$} \\
    & & 25 & 57.65±1.57 & 79.71±2.88$^2$ & 52.09±3.97$^6$ & 64.55±2.46$^4$ & 66.23±2.04$^3$ & 56.70±2.75$^5$ & \textbf{87.01±0.98$^1$} \\
    \cmidrule{2-10}
    &\multirow{6}{*}{PGD}  
     & 0 & 93.64±0.25 & 93.21±0.08$^4$ & 90.72±0.26$^5$ & \textbf{93.52±0.23$^1$} & 93.51±0.15$^2$ & 90.36±1.39$^6$ & 93.50±0.33$^3$ \\
    & & 5 & 85.23±0.21 & 86.72±0.12$^2$ & 86.42±0.38$^3$ & 85.88±0.11$^4$ & 85.51±0.15$^5$ & 84.11±0.57$^6$ & \textbf{88.86±0.24$^1$} \\
    & & 10 & 82.06±0.30 & 83.39±0.19$^2$ & 82.19±0.44$^5$ & 83.12±0.05$^3$ & 82.53±0.30$^4$ & 81.54±0.46$^6$ & \textbf{86.28±0.55$^1$} \\
    & & 15 & 79.56±0.70 & 80.37±0.69$^3$ & 79.40±0.27$^6$ & 81.04±0.08$^2$ & 80.25±0.29$^4$ & 79.62±0.47$^5$ & \textbf{83.91±0.62$^1$} \\
    & & 20 & 78.42±0.45 & 78.96±0.45$^5$ & 78.35±0.25$^6$ & 79.50±0.08$^2$ & 79.10±0.24$^3$ & 79.09±0.39$^4$ & \textbf{81.34±0.83$^1$} \\
    & & 25 & 76.46±1.03 & 77.78±0.04$^5$ & 76.51±0.98$^6$ & 78.33±0.42$^2$ & 77.95±0.39$^4$ & 77.98±0.28$^3$ & \textbf{79.62±0.77$^1$} \\

    \cmidrule{1-10}
    \multirow{12}{*}{Computers} 

    & \multirow{6}{*}{Metattack} 
     & 0 & 88.30±0.43 & 87.44±0.42$^4$ & 77.64±0.88$^5$ & 88.71±0.32$^2$ & 88.35±0.39$^3$ & * & \textbf{88.98±0.39$^1$} \\
    & & 5 & 81.68±1.06 & 83.90±0.50$^2$ & 73.10±0.72$^5$ & 78.22±0.55$^4$ & 82.18±0.66$^3$ & * & \textbf{85.58±0.53$^1$} \\
    & & 10 & 74.20±1.84 & \textbf{78.60±1.14$^1$} & 67.43±0.76$^5$ & 74.40±0.40$^4$ & 75.63±0.59$^3$ & * & 78.30±1.91$^2$ \\
    & & 15 & 72.10±1.57 & 78.06±1.26$^2$ & 64.78±1.10$^5$ & 71.57±0.94$^4$ & 74.21±0.55$^3$ & * & \textbf{80.58±1.05$^1$} \\
    & & 20 & 71.06±1.40 & 77.54±0.85$^2$ & 64.90±1.77$^5$ & 71.11±1.00$^4$ & 72.91±0.45$^3$ & * & \textbf{80.49±0.67$^1$} \\
    & & 25 & 69.50±1.16 & 76.02±1.52$^2$ & 63.92±1.44$^5$ & 70.28±0.70$^4$ & 71.95±0.54$^3$ & * & \textbf{79.83±1.73$^1$} \\
    \cmidrule{2-10}
    &\multirow{6}{*}{PGD}  
     & 0 & 88.30±0.43 & 87.26±0.38$^4$ & 77.64±0.88$^5$ & 88.69±0.31$^2$ & 88.34±0.35$^3$ & * & \textbf{88.98±0.39$^1$} \\
    & & 5 & 79.22±0.56 & 81.81±0.24$^2$ & 75.07±0.78$^5$ & 80.77±0.24$^3$ & 80.41±0.28$^4$ & * & \textbf{85.29±0.36$^1$} \\
    & & 10 & 74.40±0.74 & 77.60±0.65$^2$ & 71.37±1.26$^5$ & 77.38±0.18$^3$ & 75.02±0.49$^4$ & * & \textbf{82.95±0.69$^1$} \\
    & & 15 & 70.72±0.69 & 73.40±1.07$^3$ & 67.87±1.07$^5$ & 74.29±0.39$^2$ & 71.45±0.47$^4$ & * & \textbf{81.03±0.91$^1$} \\
    & & 20 & 68.58±0.89 & 70.30±1.41$^3$ & 65.90±1.21$^5$ & 71.76±0.31$^2$ & 69.79±0.54$^4$ & * & \textbf{79.07±0.86$^1$} \\
    & & 25 & 66.66±0.88 & 67.85±0.63$^4$ & 64.13±1.23$^5$ & 70.31±0.39$^2$ & 67.86±0.61$^3$ & * & \textbf{78.14±0.83$^1$} \\
     
    \bottomrule
    \end{tabular}
    \end{table*}

\subsection{Defense Performance}
In this section, we conduct a comprehensive evaluation of the node classification accuracy and execution time for various defense mechanisms against non-targeted attacks. Our Talos is compared with several state-of-the-art defense methods, including GCN-SVD, GCN-Jaccard, RobustGCN \citep{RGCN}, GNNGuard, and Pro-GNN. 

Specifically, in the case of the Cora, we meticulously followed the hyperparameter configurations outlined in the respective baseline defense methodologies. For the Photo and Computers, since the original defense algorithms did not have experimental results, we adjusted the hyperparameters for each method to ensure the fairness of the experiment. The precise hyperparameters are as follows: for GCN-SVD, $k=100$; for GCN-Jaccard, threshold $=0.1$; for RobustGCN, $\gamma = 0.2$. GNNGuard and Pro-GNN utilized their default settings. For the hyperparameters of Talos, validation sets were used for optimization. In addition, in order to ensure the fairness of the experiment, all the above algorithms adopt a two-layer GNN during the experiment, and the number of hidden units is 16.
% 这是pgd的表

\subsubsection{Accuracy Evaluation}
%我们首先评估了不同防御方法对非目标对抗攻击的节点分类精度。采用了上一节中使用metattack和pgd生成的对抗攻击数据集进行实验。我们分别在表1和表2中展示了两种攻击方法下带有标准差的平均准确度。最佳性能用粗体突出显示，次佳性能用斜体突出显示。从表中可以看出：
% 1. 我们的方法在不同的扰动率下基本保持领先，在高比例的扰动下效果显著。随着攻击强度的增加，各种防御方法的分类准确率都有不同程度的下降。值得注意的是，在较高的攻击强度下，Talos往往保持较高的分类准确率，显示出较强的鲁棒性。
% 2. 在干净图及低比例的攻击下，并未对GCN精度带来太大的损失。通过图结构净化的方式（例如GCN-Jaccard）进行防御的情况下往往会损害原图结构，反而在干净图中起到适得其反的作用。Talos通过累加的方式较为精细的区分不相似边从而对被攻击边有更精细的甄别能力。

We initially assess the accuracy of node classification for various defense methods against non-target adversarial attacks.   
The adversarial attack dataset, generated in a prior section using Metattack and PGD, was employed for our experiments.   
Tables~\ref{Performance under Metattack} present the mean accuracy along with the standard deviation under the two respective attack methods.   
Performance rankings are indicated in the upper right corner, with the top-performing results highlighted in bold.   
As observed from the tables, our approach consistently leads under diverse disturbance rates, demonstrating significant efficacy under higher levels of disturbance. Despite the decmidrule in classification accuracy across all defense methods as the attack intensity increased, it is worth noting that Talos tended to maintain a high classification accuracy and showed strong robustness under higher attack intensity.  
      
In scenarios involving origin graphics and low percentage attacks, GCN's accuracy is not substantially compromised.  Defense strategies that employ structure purification (such as GCN-Jaccard) often damage the original structure and have the opposite effect on graphs.   
Through cumulative means, Talos can more accurately differentiate dissimilar edges, thus better filtering out attacked edges.  
% \end{enumerate}  

\begin{table}[ht]
    \centering
    \footnotesize
    \caption{Execution time(s)}
    \vspace{10pt}
    \setlength{\tabcolsep}{3pt}
    \label{Execution time(s)}
    \begin{tabular}{cccc}
    \toprule
    DataSet & Cora & Photo & Computers \\
    \midrule
    GCN & 0.63±0.14 & 2.67±0.37 & 8.12±1.90\\
    GCN-Jaccard & 1.28±0.31 & 3.67±0.47 & 8.79±1.24 \\
    GCN-SVD & 3.42±0.01 & 46.86±0.66 & 328.82±6.62   \\
    RobustGCN & 5.66±0.02 & 8.13±0.04 & 37.71±0.06\\
    GNNGuard & 46.98±0.58 & 1239.13±148.32 & 1346.29±240.29\\
    Pro-GNN & 973.90±100.42 & 16289.10±204.53 & Time Out \\
    Talos & 0.60±0.10 & 2.80±0.20 & 11.90±1.54  \\
    \bottomrule
    \end{tabular}
    \end{table}
    
\begin{table*}[ht]
    \centering
    \footnotesize
    \caption{Universality Performance under PGD(\%)}
    \vspace{10pt}
    \label{universal under Metattack}
    \begin{tabular}{cc|c|c|ccccc}
    \toprule
     DataSet & GNN Model & Defense Method & Clean Graph & Ptb5\% & Ptb10\% & Ptb15\% & Ptb20\% & Ptb25\% \\
    \midrule
    \multirow{9}{*}{Cora} & \multirow{3}{*}{GCN} & NoDefense & \textbf{83.10±0.44} & 78.57±0.21 & 76.98±0.29 & 75.21±0.34 & 73.58±0.19 & 72.23±0.31 \\
     &  & GNNJaccard & 81.53±0.33 & \textbf{78.74±0.67} & 76.75±0.85 & 75.16±0.99 & 73.49±0.65 & 72.86±0.66 \\
     &  & Talos & 79.71±0.51 & 78.67±0.42 & \textbf{78.08±0.63} & \textbf{77.31±0.43} & \textbf{74.99±0.66} & \textbf{75.46±0.48} \\
    \cmidrule{2-9}
     & \multirow{3}{*}{GAT} & NoDefense & \textbf{83.78±0.84} & \textbf{80.31±0.61} & 77.61±0.43 & 75.54±0.74 & 73.83±0.90 & 72.40±0.75 \\
     &  & GNNJaccard & 82.13±0.78 & 80.16±1.18 & 78.15±1.36 & 76.70±1.08 & 74.48±1.06 & 73.63±1.31 \\
     &  & Talos & 83.10±1.20 & 79.85±1.01 & \textbf{79.71±1.37} & \textbf{78.79±0.82} & \textbf{77.55±0.95} & \textbf{76.45±0.94} \\
    \cmidrule{2-9}
     & \multirow{3}{*}{GragphSAGE} & NoDefense & \textbf{82.98±0.50} & 79.68±0.60 & 77.55±0.50 & 76.35±0.65 & 74.36±0.61 & 73.49±0.38 \\
     &  & GNNJaccard & 81.52±0.51 & 80.18±0.55 & 78.59±0.40 & 77.21±0.62 & 75.55±0.64 & 75.16±0.50 \\
     &  & Talos & 82.96±0.39 & \textbf{80.42±0.44} & \textbf{79.10±0.83} & \textbf{78.72±0.47} & \textbf{77.68±0.35} & \textbf{77.17±0.48} \\
    \cmidrule{1-9} \cmidrule{2-9}
    \multirow{9}{*}{Photo} & \multirow{3}{*}{GCN} & NoDefense & \textbf{93.88±0.18} & 85.27±0.26 & 82.18±0.30 & 80.04±0.35 & 78.79±0.47 & 77.72±0.58 \\
     &  & GNNJaccard & 93.29±0.16 & 87.17±0.31 & 83.75±0.40 & 80.81±0.26 & 79.28±0.25 & 77.91±0.43 \\
     &  & Talos & 93.62±0.19 & \textbf{89.73±0.20} & \textbf{88.56±0.35} & \textbf{86.68±0.32} & \textbf{86.37±0.49} & \textbf{84.53±0.64} \\
    \cmidrule{2-9}
     & \multirow{3}{*}{GAT} & NoDefense & \textbf{94.10±0.22} & 85.02±0.21 & 82.64±0.44 & 80.38±0.59 & 78.77±1.26 & 77.89±0.37 \\
     &  & GNNJaccard & 93.70±0.20 & 87.09±0.34 & 83.92±0.18 & 81.87±0.34 & 80.13±0.44 & 79.06±0.35 \\
     &  & Talos & 93.99±0.25 & \textbf{90.20±0.29} & \textbf{88.98±0.37} & \textbf{87.27±0.29} & \textbf{86.49±0.39} & \textbf{85.99±0.51} \\
    \cmidrule{2-9}
     & \multirow{3}{*}{GragphSAGE} & NoDefense & \textbf{94.41±0.21} & 88.06±0.25 & 85.19±0.49 & 83.78±0.49 & 82.61±0.45 & 82.03±0.48 \\
     &  & GNNJaccard & 94.18±0.27 & 90.32±0.28 & 87.64±0.26 & 85.94±0.46 & 84.44±0.33 & 82.89±0.68 \\
     &  & Talos & 94.16±0.19 & \textbf{91.43±0.21} & \textbf{90.54±0.33} & \textbf{89.26±0.56} & \textbf{88.89±0.34} & \textbf{88.50±0.30} \\
    \cmidrule{1-9} \cmidrule{2-9}
    \multirow{9}{*}{Computers} & \multirow{3}{*}{GCN} & NoDefense & 88.80±0.24 & 80.68±0.47 & 74.24±1.13 & 70.42±1.05 & 67.86±1.06 & 65.96±1.01 \\
     &  & GNNJaccard & 88.36±0.39 & 82.76±0.44 & 79.15±0.51 & 75.00±1.53 & 70.57±1.49 & 68.30±1.14 \\
     &  & Talos & \textbf{88.98±0.39} & \textbf{85.29±0.36} & \textbf{82.95±0.69} & \textbf{81.03±0.91} & \textbf{79.07±0.86} & \textbf{78.14±0.83} \\
    \cmidrule{2-9}
     & \multirow{3}{*}{GAT} & NoDefense & 89.22±0.19 & 80.89±0.46 & 77.05±1.61 & 71.65±2.58 & 69.82±2.50 & 66.74±2.85 \\
     &  & GNNJaccard & 88.24±0.45 & 82.76±0.47 & 78.80±1.07 & 74.54±1.90 & 73.19±1.76 & 69.10±2.29 \\
     &  & Talos & \textbf{89.41±0.48} & \textbf{85.22±0.42} & \textbf{83.66±0.58} & \textbf{82.58±0.63} & \textbf{81.72±0.48} & \textbf{81.46±0.50} \\
    \cmidrule{2-9}
     & \multirow{3}{*}{GragphSAGE} & NoDefense & \textbf{88.50±0.71} & 81.96±1.24 & 78.10±1.33 & 75.66±1.68 & 74.07±1.49 & 71.36±1.41 \\
     &  & GNNJaccard & 87.10±0.66 & 83.60±0.84 & 81.03±0.77 & 79.02±1.40 & 77.24±0.98 & 75.36±2.01 \\
     &  & Talos & 87.63±0.55 & \textbf{84.99±0.99} & \textbf{83.07±1.35} & \textbf{81.53±1.78} & \textbf{80.88±0.99} & \textbf{81.19±1.21} \\
    \bottomrule
    \end{tabular}
    \end{table*}

\subsubsection{Execution Time }
Execution time is a crucial metric for evaluating defense methods. However, comparing their practicality based solely on time complexity is challenging due to various parallel computing techniques and defensive strategies. For example, GCN-Jaccard focuses on graph structure purification and doesn't include GNN fitting processes, while methods like RobustGCN do. To demonstrate practicality accurately, we measure execution time from the start of each retraining session to the completion of model fitting. Across six experiments with attack ratios from 0 to 25\%, each group was conducted ten times, yielding an average running time recorded in Table~\ref{Execution time(s)}.

To ensure fairness, both graph structure purification methods (GCN-Jaccard, GCN-SVD, and Talos) and methods employing robust graph neural networks (RobustGCN, GNNGuard, Pro-GNN) utilized similar configurations, including a two-layer GCN with 16 hidden units and a learning rate of 0.01. Talos operates significantly faster than almost any other defense method, as shown in the table. Although the experiment was conducted on a cloud computing platform where concurrent GPU usage may have influenced execution times, Talos closely resembles the processing speed of GCN, indicating its efficiency. Further investigation revealed that the average operation times for Talos across different datasets were 0.040s, 0.378s, and 2.582s for Cora, Photo, and Computers, respectively. Given its faster processing time compared to GCN, Talos emerges as an effective and practical strategy for purifying GNN data before defense.

\subsubsection{Larger Scale Graph Application}
%这里写Pubmed的部分
% 由于实验条件的限制，我们无法对Pubmed进行Metattack的攻击，同时后续的防御实验也无法进行。为了验证Talos是否在更大的图上有较好的防御效果，我们依然选择在Pubmed上进行实验。因此我们参考了 Daniel Zügner~\citep{SoftMedian} 等人的做法，采用PR-BCD对Pubmed的图结构分别进行了1\%,5\%和10\%的扰动攻击。 同时我们与Vanilla GCN，Vanilla GDC，GCNJaccard，SoftMedian等Baseline进行了效果对比，为了保证实验的公平性，我们都采用了同样的GCN结构，隐藏单元数为64。对于SoftMedian我们则采用了SoftMedian GDC（T=1）的解决方案，同时也尝试将Talos替换GDC对数据的预处理，与SoftMedian结合使用。精度和时间的实验结果如表~\ref{table:pubmed}中所示。

\begin{figure}[ht]
    \centering
    \includegraphics[width=0.95\linewidth]{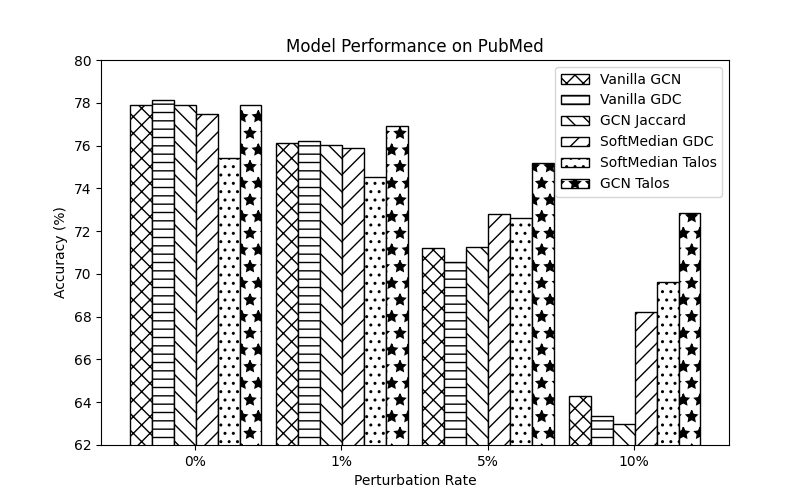}
    \caption{Model Performance on Pubmed}
    \vspace{10pt}
    \label{fig:pubmed}
\end{figure}

%通过实验可以看出，Talos在PR-BCD的扰动攻击下在大型图下仍能保持最好的防御效果，且横向对比下用时最少。
Due to the limitations of the experimental platform, we were unable to perform Metattack attacks on Pubmed, and subsequent defense experiments were also not feasible. To verify whether Talos has a better defensive effect on larger graphs, we still chose to conduct experiments on Pubmed. Therefore, we referred to the approach of Daniel Zügner~\citep{SoftMedian} and used PR-BCD to perturb the graph structure of Pubmed with 1\%, 5\%, and 10\% attacks, respectively. We also compared the effects with Baselines such as Vanilla GCN, Vanilla GDC, GCNJaccard, and SoftMedian. To ensure fairness in the experiments, we all used the same GCN structure with 64 hidden units. For SoftMedian, we adopted the SoftMedian GDC (T=1) solution and also tried replacing GDC with Talos for data preprocessing, combining its use with SoftMedian. The accuracy is shown in Figure~\ref{fig:pubmed}.

As for the time taken for the experiments, according to the records, Vanilla GCN and Vanilla GDC took 1.63s and 5.80s respectively. With defenses applied, GCNJaccard, SoftMedian GDC, SoftMedian Talos, and GCN Talos took 14.24s, 21.29s, 16.92s, and 11.61s, respectively. It can be observed that GCN Talos has the shortest overall time consumption.

The experiments show that Talos maintains the best defensive effect under PR-BCD perturbation attacks on large graphs and has the least time consumption in a lateral comparison.

\subsection{Universality of Talos Over GNNs}
Talos only uses the information of the graph itself and the assumption of homophily property.  Therefore, Talos is theoretically applicable to all kinds of GNNs.  Although previous experiments have been tested on GCN to verify its effectiveness against other defense methods, it is unclear whether it is also better on other GNNs.  In order to verify the generality of Talos on GNNs, this section selects several common GNNs for testing, including GCN, GAT, and GraphSage.  The experimental results are shown in Table~\ref{universal under Metattack}. Due to the large scale of the Computers dataset, on the experimental platform, Pro-GNN failed to complete the computation and produce results within the predetermined time frame. Therefore, we marked it as a timeout and excluded Pro-GNN from the comparative analysis of the Computers.

As can be seen from the Table, Talos achieves significant results in all three basic models, so Talos essentially purifies the graph structure regardless of the selected model.  The defense of Talos over GNNs is universal.

\subsection{Why is Talos Effective}
%在先前的实验中，我们发现Talos在处理图结构数据时显著地展现出净化能力。为了探讨Talos有效性的本质，我们对其与GCNJaccard方法进行了深入对比分析。具体而言，我们采用了PGD Attack在25%攻击比例下产生的干扰图，并通过绘制在GCNJaccard和Talos视角下，攻击边与干净边的分布直方图，以深入分析两者对扰动边的鉴别能力。

\begin{figure}[ht]
    \centering
    \includegraphics[width=0.95\linewidth]{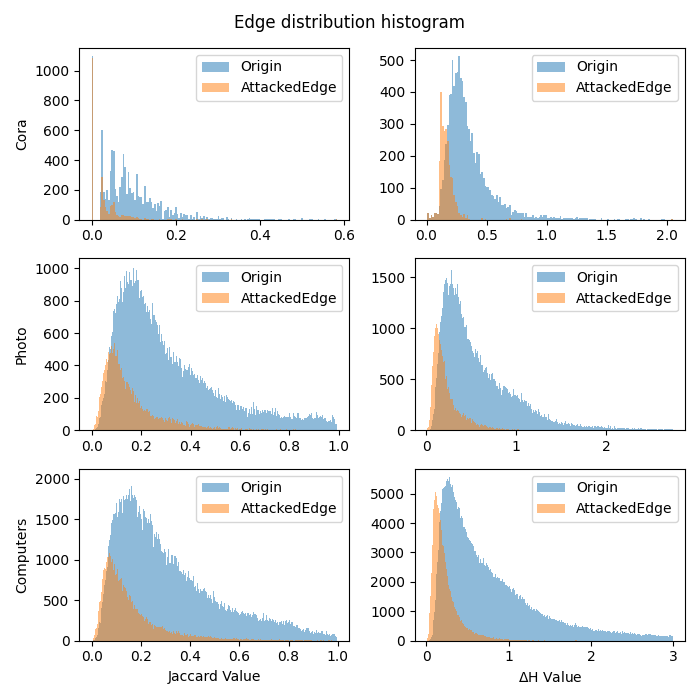}
    \caption{Edge distribution histogram under PGD}
    \vspace{10pt}
    \label{fig:histogram}
        % \vspace{-5pt}

\end{figure}

To explore why Talos is effective, we conducted a comparative analysis between Talos and GCN-Jaccard. Using the PGD Attack, we generated attacked graphs with a 25\% attack rate. By plotting histograms of attack and clean edges for graphs processed by GCN-Jaccard and Talos, we compared their abilities to distinguish perturbed edges. Figure~\ref{fig:histogram} illustrates that Talos, with its $\Delta \mathrm{Hom}$ metric, widens the distribution gap between attack and clean edges compared to GCN-Jaccard. It also creates a clustering effect on attack edges, facilitating their identification and removal.

% 我们还采用了两个量化指标来计算攻击边和干净边的分布差异。
% 首先，我们采用Kolmogorov-Smirnov（KS）指标对两个分布进行量化。KS指标是统计学中用于衡量两个分布之间差异的一个量。KS指标的范围从0到1，KS值越大，表示模型的区分能力越强。一般认为，KS指标在0.2到0.5之间，模型的区分能力良好；KS指标在0.5以上，模型的区分能力非常强。
% 经过计算，对于Jaccard，Ks cora = 0.46，ks photo=0.39，ks computers=0.38。而对于Talos的dH，Ks cora = 0.66，ks photo=0.51，ks computers=0.57。可以看出Talos对于攻击边的区分能力非常强，并且远超Jaccard。
% 其次，我们定义计算了对于两者而言，攻击边的分离率：
We also adopted two quantitative metrics to measure the distribution difference between attack edges and clean edges. Firstly, we used the Kolmogorov-Smirnov ($\mathbf{KS}$) metric to quantify the difference between the two distributions. The KS metric is a measure in statistics used to gauge the disparity between two distributions. The range of the KS metric is from 0 to 1, with a higher KS value indicating a stronger discriminative capability of the model. Generally, a KS metric between 0.2 and 0.5 is considered to have a good discriminative capability; above 0.5, the model's discriminative capability is very strong.
After calculation, for Jaccard, $\mathbf{KS}_{\textbf{cora}} = 0.46$, $\mathbf{KS}_{\textbf{photo}} = 0.39$, $\mathbf{KS}_{\textbf{computers}} = 0.38$. For Talos' $ \Delta  \mathrm{Hom}$, $\mathbf{KS}_{\textbf{cora}} = 0.66$, $\mathbf{KS}_{\textbf{photo}} = 0.51$, $\mathbf{KS}_{\textbf{computers}} = 0.57$. The results indicate that Talos has a very strong discriminative capability for attack edges and significantly surpasses Jaccard.

Secondly, we defined and calculated the separation rate ($\mathbf{SR}$) for attack edges for both by 
$\mathbf{SR} = \frac{S_{\textbf{attacked}}-S_{\textbf{overlap}}}{S_{\textbf{attacked}}}$
where $S_{\textbf{attacked}}$ is histogram area of attack edge distribution, $S_{\textbf{overlap}}$is area of overlap between the two histograms.
After calculation, for Jaccard, $\mathbf{SR}_{\textbf{cora}} = 0$, $\mathbf{SR}_{\textbf{photo}} = 6.93\%$, $\mathbf{SR}_{\textbf{computers}} = 6.45\%$. For Talos' $ \Delta  \mathrm{Hom}$, $\mathbf{SR}_{\textbf{cora}} = 29.8\%$, $\mathbf{SR}_{\textbf{photo}} = 22.3\%$, $\mathbf{SR}_{\textbf{computers}} = 27.5\%$. It can be seen that Talos has a very strong discriminative capability for attack edges and significantly surpasses Jaccard.

% 综上所述，Talos之所以有更好的效果，其本质是对攻击边和干净边的分布有更好的区分能力，防御者在进行图净化的过程中，可以在减少对原图造成损害的同时净化更多的攻击边，从而达到更好的防御效果。
In summary, the reason why Talos has a better effect is its essence in better distinguishing the distribution between attack edges and clean edges. During the graph purification process, defenders can purify more attack edges while minimizing damage to the original graph, thus achieving better defensive results.

\section{Conclusion}
This paper introduces Talos, a generalized defense method for improving the robustness of GNN models. Talos enhances the global homophily of graphs, thus removing adversarial edges to a greater extent. We show that with proper approximations, Talos can be computed very efficiently. Comparative experiments validate Talos' advantages in terms of time and accuracy, confirming its universality across different GNN models.

%%%%%%%%%%%%%%%%%%%%%%%%%%%%%%%table%%%%%%%%%%%%%%%%%%%%%%%%%%%%%%%%%%%%%%%

%%%%%%%%%%%%%%%%%%%%%%%%%%%%%%%table%%%%%%%%%%%%%%%%%%%%%%%%%%%%%%%%%%%%%%%

% \section{Citations and references}

%%%%%%%%%%%%%%%%%%%%%%%%%%%%%%%%%%%%%%%%%%%%%%%%%%%%%%%%%%%%%%%%%%%%%%%%

%%% Use this environment to include acknowledgements (optional).
%%% This will be omitted in doubleblind mode.

\begin{ack}
This project is supported by the National Natural Science Foundation of China under grant No. 62306328 and the science and technology innovation Program of Hunan Province under grant No. 2023RC3021. 
\end{ack}

%%%%%%%%%%%%%%%%%%%%%%%%%%%%%%%%%%%%%%%%%%%%%%%%%%%%%%%%%%%%%%%%%%%%%%%%

%%% Use this command to include your bibliography file.

\bibliography{main}

\appendix
\onecolumn

\newpage

\section{Hyperparameter Analysis}

In this study, Talos has two hyperparameters: the number of edges removed ($n_r$) and the rate of information loss ($\alpha$).

\subsection{Number of Edges Removed ($n_r$)}
The choice of the number of edges removed is a critical hyperparameter. Removing too few edges may lead to poor defense performance, while removing too many may cause over-purification issues. Recent studies \cite{mao2023demystifyingstructuraldisparitygraph} indicate that GNNs rely on the graph's mainstream structure, whereas adversarial perturbations are non-mainstream. Therefore, in clean graphs, slightly removing weakly homophilic edges shows little negative impact. However, if Talos exhausts heterophilic edges and continues removing homophilic parts, over-purification may occur. We addressed this issue using a validation set. Over-purification increases validation error by disrupting the mainstream structure. This signal helps us halt the purification process at a reasonable stage.

\subsection{Rate of Information Loss ($\alpha$)}
For $\alpha$, according to Equation \ref{equ:3}, it should be within the range $[0, \frac{1}{\rho(\mathbf{A})}]$ for Talos to converge. Within this range, Talos performs stably, except near 0. Therefore, selecting a value slightly below $\frac{1}{\rho(\mathbf{A})}$ yields good results.

Regarding the calculation of $\rho(\mathbf{A})$, we know that computing $\rho(\mathbf{A})$ is resource-intensive for large graphs. Guo et al. \cite{pubanjing} show that the spectral radius of a graph satisfies the following formula:

\[
\rho(\mathbf{A}) \leq \sqrt{2 |\mathcal{E}| - |\mathcal{V}| - k + \frac{5}{2} + \sqrt{2 |\mathcal{E}| - 2 |\mathcal{V}| + \frac{9}{4}}}
\]

where \(3 \leq k \leq n\) is an integer that satisfies the following condition:

\[
\frac{(k-2)(k-3)}{2} \leq m - n \leq \frac{k(k-3)}{2}
\]

For small graphs, it is feasible to calculate the exact value of $\rho(\mathbf{A})$. However, for larger graphs, we can use the above formula to estimate $\rho(\mathbf{A})$ and thus select the hyperparameters accordingly.

\section{Comparing Mid-GCN and Talos Defense}
\setcounter{table}{0}
\renewcommand{\thetable}{B\arabic{table}}
\begin{table}
\centering
\label{tab:midgcn}
\begin{tabular}{ccc|c|ccccc|c}
\toprule
DataSet & Defense Method & Attack & Clean Graph & Ptb5\% & Ptb10\% & Ptb15\% & Ptb20\% & Ptb25\% & Time \\

\midrule

\multirow{4}{*}{Cora} & \multirow{2}{*}{Mid-GCN} & Metattack & 82.31$\pm$0.47 & 79.80$\pm$0.61 & 76.03$\pm$0.71 & 69.51$\pm$1.09 & 60.48$\pm$1.51 & 59.11$\pm$1.86 &   \multirow{2}{*}{0.73$\pm$0.10 }\\
     &         & PGD       & 82.28$\pm$0.53 & \textbf{78.88$\pm$0.49} & 76.29$\pm$0.42 & 74.27$\pm$0.57 & 72.79$\pm$0.66 & 71.37$\pm$0.56 &            \\
     \cmidrule{2-10}
&  \multirow{2}{*}{Talos}     & Metattack & 82.71$\pm$0.49 & \textbf{80.83$\pm$0.74 }& \textbf{78.31$\pm$0.76 }& \textbf{74.17$\pm$0.39 }& \textbf{68.19$\pm$0.98 }& \textbf{65.28$\pm$1.16} &  \multirow{2}{*}{0.60$\pm$0.10}             \\
     &         & PGD       & 81.06$\pm$0.91 & 78.81$\pm$0.47 & \textbf{78.42$\pm$0.54 }& \textbf{75.97$\pm$0.41 }& \textbf{75.40$\pm$0.64} & \textbf{73.98$\pm$0.34} &             \\
\midrule
\multirow{4}{*}{Photo}  & \multirow{2}{*}{Mid-GCN} & Metattack & 92.05$\pm$0.23 & 88.26$\pm$0.49 & 85.87$\pm$0.64 & 76.35$\pm$3.55 & 72.97$\pm$2.88 & 71.19$\pm$2.86 & \multirow{2}{*}{4.15$\pm$0.22} \\
      &         & PGD       & 91.92$\pm$0.27 & 84.45$\pm$0.59 & 81.80$\pm$0.57 & 79.64$\pm$0.89 & 78.05$\pm$1.50 & 76.79$\pm$1.75 &             \\
     \cmidrule{2-10}
      
& \multirow{2}{*}{Talos}      & Metattack & 93.60$\pm$0.27 &\textbf{ 92.17$\pm$0.38} & \textbf{90.34$\pm$0.49 }& \textbf{88.60$\pm$0.70} &\textbf{ 87.29$\pm$0.66 }& \textbf{87.01$\pm$0.98 }&  \multirow{2}{*}{12.83$\pm$2.75}             \\
      &         & PGD       & 93.50$\pm$0.33 & \textbf{88.86$\pm$0.24} &\textbf{ 86.28$\pm$0.55 }&\textbf{ 83.91$\pm$0.62} & \textbf{81.34$\pm$0.83 }&\textbf{ 2.80$\pm$0.20} &             \\
\midrule
\multirow{4}{*}{Computers}  &  \multirow{2}{*}{Mid-GCN} & Metattack & 85.62$\pm$0.34 & 80.02$\pm$0.47 & 73.47$\pm$0.89 & 72.19$\pm$0.93 & 70.67$\pm$0.96 & 69.81$\pm$1.37 & \multirow{2}{*}{12.83$\pm$2.75} \\
          &         & PGD       & 85.64$\pm$0.32 & 77.25$\pm$0.85 & 72.40$\pm$0.71 & 69.33$\pm$0.83 & 67.58$\pm$0.82 & 65.75$\pm$1.19 &             \\
     \cmidrule{2-10}

          & \multirow{2}{*}{Talos}      & Metattack & 88.98$\pm$0.39 & \textbf{85.58$\pm$0.53} & \textbf{78.30$\pm$1.91 }& \textbf{80.58$\pm$1.05} & \textbf{80.49$\pm$0.67 }& \textbf{79.83$\pm$1.73 }&  \multirow{2}{*}{12.83$\pm$2.75}             \\
          &         & PGD       & 88.98$\pm$0.39 &\textbf{ 85.29$\pm$0.36 }& \textbf{82.95$\pm$0.69} & \textbf{81.03$\pm$0.91 }& \textbf{79.07$\pm$0.86 }&\textbf{ 11.90$\pm$1.54 }&             \\
\bottomrule
\end{tabular}
\caption{Performance Comparison of Mid-GCN and Talos under Adversarial Attacks}
\vspace{10pt}
\end{table}

% We supplemented our study with experiments evaluating the effectiveness of two defense methods, Mid-GCN and Talos, against various attacks across different datasets, including Cora, Photo, and Computers. The attacks examined include Metattack and PGD, with results presented in Table \ref{tab:midgcn}. The findings indicate that while Mid-GCN performs well in certain settings, Talos provides superior protection in most high-perturbation scenarios. Notably, Talos demonstrates more robust and efficient defense capabilities under conditions of high perturbation.
In recent research advancements, the Mid-GCN\cite{huang2023robust} model has been demonstrated as an effective method for countering adversarial attacks on graph neural networks (GNNs). Traditional defense strategies often rely on restoring the original graph structure or imposing robustness constraints during training to resist attacks. However, Mid-GCN employs an innovative approach by leveraging mid-frequency signals to enhance the model's resilience against adversarial manipulations. Specifically, Mid-GCN integrates band-pass filters, allowing it to selectively emphasize graph signals that are less susceptible to noise and adversarial distortions—signals that are often overlooked by other methods.

To further evaluate the efficacy of Mid-GCN, we conducted comparative experiments with Talos, assessing their performance against various attacks, including Metattack and PGD, across multiple datasets such as Cora, Photo, and Computers. The experimental results, presented in Table \ref{tab:midgcn}, indicate that while Mid-GCN performs well in certain settings, Talos demonstrates superior protective capabilities in most high-disturbance scenarios. Particularly in high-perturbation conditions, Talos not only provides a more robust defense mechanism but also exhibits faster response times. This suggests that in complex attack environments, Talos is more effective and swift in safeguarding graph structures from adversarial influences, thereby maintaining the stability and accuracy of the model.

% \section{Limitations and Future Research Directions of Talos}

\section{Limitations}

Despite the excellent performance of Talos in utilizing global homophily for adversarial defense in GNNs, its application scope has certain limitations. Specifically, Talos may not be effective when dealing with graphs exhibiting weak homophily. This limitation arises because the defense mechanism of Talos relies on the global homophily of the graph, which is insufficient to distinguish between normal and adversarial edges in graphs with weak homophily. To avoid the inappropriate use of Talos, we recommend measuring the homophily of a graph before applying Talos to ensure the effectiveness of the defense mechanism.

\end{document}